\documentclass[mlabstract]{jmlr}

\usepackage[hpos=300px,vpos=70px]{draftwatermark}
\SetWatermarkText{\test}
\SetWatermarkScale{1}
\SetWatermarkAngle{0}

\usepackage{longtable}
\usepackage{float}
\usepackage{graphicx}

\usepackage{booktabs}
\usepackage[load-configurations=version-1]{siunitx} 

\theorembodyfont{\upshape}
\theoremheaderfont{\scshape}
\theorempostheader{:}
\theoremsep{\newline}

\jmlrvolume{}
\firstpageno{1}
\editors{List of editors' names}

\jmlryear{2026}
\jmlrworkshop{Symmetry and Geometry in Neural Representations}
\newcommand\norm[1]{\left\lVert#1\right\rVert}

\newif\ifanonymous
\anonymousfalse
\title[Short Title]{Parameter-Level Attribution of Symmetry in Trained Networks Though Parameter-Wise Functional Sensitivity}

\ifanonymous
\author{%
  \Name{Anonymous Author(s)}\\
  \addr Anonymous Institution
}
\else
\author{%
  \Name{Alan Muriithi} \Email{muriithi@maths.ox.ac.uk}\\
  \Name{Vedanta Thapar} \Email{vedanta.thapar@maths.ox.ac.uk}\\
  \addr Mathematical Institute, University of Oxford
  \AND
  \Name{Torben Berndt} 
  \Email{}\\
  \addr Heidelberg Institute for Theoretical Studies
}
\fi

\begin{document}

\maketitle

\begin{abstract}
When a network has learned a function with a known symmetry, can that symmetry be moved through the parametrisation---is there a motion in parameter space realising the group action in function space? We formulate this as a lifting problem for the realisation map $\Phi:\theta\mapsto f_\theta$, and show that a smooth parameter-space action exists only if the tangent space to the function's symmetry orbit lies within the image of $\mathrm d\Phi_\theta$, whose columns are the \emph{functional sensitivities} of individual parameters. This condition is also sufficient for pointwise first-order
lifting. Relaxing it in least squares yields two local parameter directions: one following the symmetry orbit, one descending towards the equivariant subspace, with residuals measuring what the parametrisation cannot reach. On a rotationally invariant classifier we find these directions induce their predicted function-space motion, but only locally: recomputed directions track the orbit and reduce the equivariance defect, while directions held fixed depart from both after training. The same holds for Hamiltonian neural networks trained on a rotationally symmetric potential, even though the architecture does not explicitly enforce the symmetry.
\end{abstract}


\section{Introduction}
\label{sec:Introduction}

A fundamental question in the study of parameterised function classes is to identify the subsets of parameters that play a dominant role in the behaviour of the learned map. We approach this by defining the \textit{functional sensitivity} of a neural network, which measures the local response of the output map to small perturbations of a parameter. Concretely, consider a map $f_{\theta} : \mathcal{X} \to \mathcal{Y}$, parameterised by a set of weights and biases encoded in $\theta$. The pointwise functional sensitivity of the $i$-th parameter, $\theta_i$, at an input $x\in \mathcal{X}$ is defined as
\[ S^*_i(x;\theta) := \frac{\partial f_{\theta}}{\partial \theta_i}(x;\theta) \in \mathbb{R}^{\dim\mathcal{Y}},\]
and the aggregate functional sensitivity is obtained by averaging its magnitude over inputs,
\[S_i(\theta) := \Big(\tfrac{1}{\dim\mathcal Y}\,\mathbb{E}_{x\sim \mathcal{D}}\big[\lVert S_i^*(x;\theta)\rVert_2^{2}\big]\Big)^{1/2}, \]
where $\mathcal{D}$ is a probe distribution on $\mathcal{X}$; the discretisation, and the sensitivity-normalised gauge in which all per-parameter attributions below are computed, are given in Appendix~\ref{app:numerics}.
Mathematically, this object is related to the neural tangent kernel (NTK) \citep{Jacot_2018}, which considers the inner product between pointwise sensitivities at different inputs for fixed $\theta$. Where the NTK is a tool for the asymptotic analysis of optimisation in the infinite-width limit, the functional sensitivity shifts the focus from comparing the response of the map at different inputs to comparing the roles of individual parameters.

We study functional sensitivity in networks trained from sampled trajectory data, aiming to identify which parameters encode symmetries of the learned map. Our main contributions are:

(i) we introduce \emph{functional sensitivity}, a parameter-level differential framework relating individual parameters to local variations of the learned map; (ii) we develop a projection of infinitesimal symmetry generators onto the parameter tangent space, yielding per-parameter attributions of symmetry and a diagnostic for equivariance; and (iii) we empirically compare these quantities in Hamiltonian neural networks, identifying orthogonal directions along the symmetry orbit and towards the equivariant subspace.

\citet{Maiti_2021} infer the manifestation of symmetries in the learned function through parameter-space correlators. In contrast, our method analyses the local differential geometry of the learned map rather than statistical correlations in parameter space. Furthermore, our approach provides parameter-level attribution of symmetry generators, enabling the identification of the specific parameters that contribute to a given symmetry.


\section{Symmetry Orbits and Infinitesimal Lifting}
\label{sec:symmetry-realisability}

This section studies when a symmetry action on function space can be lifted through the parametrisation of a neural network. Figure~\ref{fig:conceptual-overview} illustrates the construction; a more rigorous treatment is given in Appendix~\ref{app:symmetry-realisability}.

\subsection{Setup and the Differential Lifting Criterion}
\label{sec:infinitesimal-lifting}

Let \(G\) be a finite-dimensional Lie group acting smoothly on \(\mathcal X\) by \(\tau:G\times\mathcal X\to\mathcal X\), and let \((\rho,V)\) be a finite-dimensional representation of \(G\). On a hypothesis space \(\mathcal B\) of differentiable maps \(f:\mathcal X\to V\), these actions induce
\(\Pi:G\to\operatorname{GL}(\mathcal B)\) with
\([\Pi(g)f](x)=\rho(g)f(\tau_{g^{-1}}x)\); its fixed points are precisely the equivariant functions. A neural network \(f_\theta\in\mathcal B\) is specified by parameters \(\theta\in\Theta\), and we denote the realisation map by
\(\Phi:\Theta\to\mathcal B\), \(\theta\mapsto f_\theta\), with realised model class \(\mathcal M:=\Phi(\Theta)\).

While \(g\in G\) acts naturally on \(f_\theta\) in function space, it is not clear whether \(\Pi(g)f_\theta\) is again realised by some parameter vector. Ideally, there exists a smooth action \(\beta:G\times\Theta\to\Theta\) such that \(\Phi(\beta_g\theta)=\Pi(g)\Phi(\theta)\). Such a global lift need not exist, so we study its weaker infinitesimal counterpart: each generator of \(G\) induces an infinitesimal direction in function space, while the network Jacobian determines which function-space directions are reachable through parameter variations, and if the symmetry action can be lifted, the former must lie within the latter.

For \(f_\theta\), let \(o_{f_\theta}(g)=\Pi(g)f_\theta\) be its orbit map and define
\(a_{f_\theta}:=\mathrm d(o_{f_\theta})_e:\mathfrak g\to T_{f_\theta}\mathcal B\).
Likewise, let
\(J_\theta:=\mathrm d\Phi_\theta:T_\theta\Theta\to T_{f_\theta}\mathcal B\)
be the differential of the realisation map, so that the locally accessible directions are
\(\mathcal T_\theta:=\operatorname{Im}(J_\theta)\).
If a global lift \(\beta\) exists, defining \(b_\theta(g):=\beta_g\theta\) gives
\(\Phi\circ b_\theta=o_{f_\theta}\); differentiating at the identity yields
\begin{equation}
    J_\theta L_\theta=a_{f_\theta},\quad L_\theta:=\mathrm d(b_\theta)_e,
    \qquad\text{whence}\qquad
    T_{f_\theta}(G\!\cdot\!f_\theta)\subseteq\operatorname{Im}(J_\theta)
    \label{eq:linear-lift-main}
\end{equation}
is necessary for lifting. At fixed \(\theta\), the inclusion in \eqref{eq:linear-lift-main} is also sufficient for pointwise first-order lifting, although not for the existence of a smooth local or global action on parameter space.

Concretely, when \(\Theta\subseteq\mathbb R^p\) the columns of \(J_\theta\) are
exactly the pointwise functional sensitivities of Section~\ref{sec:Introduction},
\(\mathcal T_\theta=\operatorname{span}\{S^*_1(\,\cdot\,;\theta),\dots,S^*_p(\,\cdot\,;\theta)\}\),
so the coefficients \(c^\star\) below are a per-parameter attribution of the symmetry direction.

\subsection{Local Alignment Objectives}
\label{sec:approximate-lifting}

We probe this local geometry in two complementary ways: whether the network can move along a symmetry direction, and whether it can move towards the equivariant subspace.

\paragraph{Symmetry directions.}
For a fixed generator \(\xi\in\mathfrak g\), let
\(\eta_\xi:=a_{f_\theta}(\xi)\) be the induced direction in function space (we reserve \(g\) for group elements).
An exact infinitesimal lift exists when \(\eta_\xi\in\operatorname{Im}(J_\theta)\). Otherwise, we take the minimum-$\ell_2$-norm parameter direction that best approximates \(\eta_\xi\),
\begin{equation}
    c^\star
    =
    J_\theta^{+}\eta_\xi
    =
    \operatorname*{arg\,min}\nolimits_{c}
    \bigl\{
        \lVert c\rVert_2 : c\in \operatorname*{arg\,min}\nolimits_{z} \lVert J_\theta z-\eta_\xi\rVert_2
    \bigr\},
    \label{eq:min_norm_attribution}
\end{equation}
whose residual \(\lVert J_\theta c^\star-\eta_\xi\rVert_2\) measures the component of the symmetry direction that is locally inaccessible. Here \(J_\theta^{+}\) is the \emph{truncated} pseudoinverse: singular values below \(\tau\sigma_{\max}(J_\theta)\), with \(\tau=10^{-3}\), are discarded, so \(\mathcal T_\theta\) is the numerically resolved tangent space rather than the full span of an overparameterised Jacobian, and the minimum-norm solve is performed in the sensitivity-normalised gauge of Appendix~\ref{app:numerics}.

\paragraph{Towards the equivariant subspace.}
The previous objective measures movement along a symmetry orbit, but not whether the model can move towards equivariance. Let
\(\mathcal E:=\mathcal B^G\) be the equivariant subspace, with orthogonal projection \(P_{\mathcal E}\), and define
\(
    \delta_\theta:=f_\theta-P_{\mathcal E}f_\theta,
\)
so that \(-\delta_\theta\) points from \(f_\theta\) towards its projection onto \(\mathcal E\). Since \(\langle\delta_\theta,\eta_\xi\rangle=0\) (Appendix~\ref{app:defect}), the two directions answer genuinely different questions: \(\eta_\xi\) is tangent to the level set of \(\lVert\delta_\theta\rVert\). We again compute the best locally accessible parameter direction,
\begin{equation}
    b^\star
    =
    J_\theta^{+}(-\delta_\theta)
    =
    \operatorname*{arg\,min}\nolimits_{c}
    \bigl\{
        \lVert c\rVert_2 : c\in \operatorname*{arg\,min}\nolimits_{z} \lVert J_\theta z+\delta_\theta\rVert_2
    \bigr\},
    \label{eq:min_norm_delta}
\end{equation}
with residual \(\lVert J_\theta b^\star+\delta_\theta\rVert_2\) measuring how much of the direction towards equivariance lies outside the locally accessible tangent space.

\section{Experiments}

We evaluate whether the local parameter-space directions introduced in Section~\ref{sec:symmetry-realisability} induce their predicted finite changes in function space.

\subsection{Annulus Classifier} 
\begin{figure}[t]
    \centering
    \begin{minipage}[t]{0.48\linewidth}
        \centering
        \includegraphics[width=\linewidth]
        {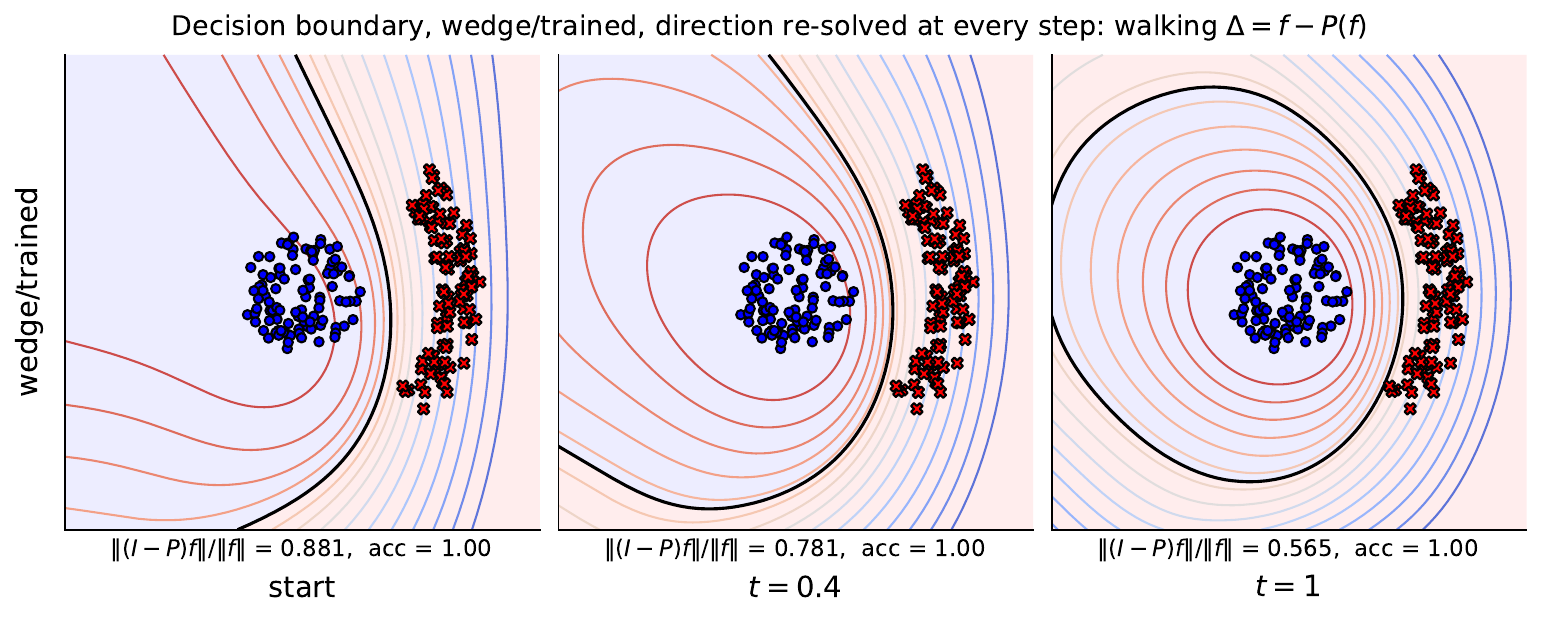}

        \smallskip
        {\small (a) Invariance-directed flow}
    \end{minipage}
    \hfill
    \begin{minipage}[t]{0.48\linewidth}
        \centering
        \includegraphics[width=\linewidth]
        {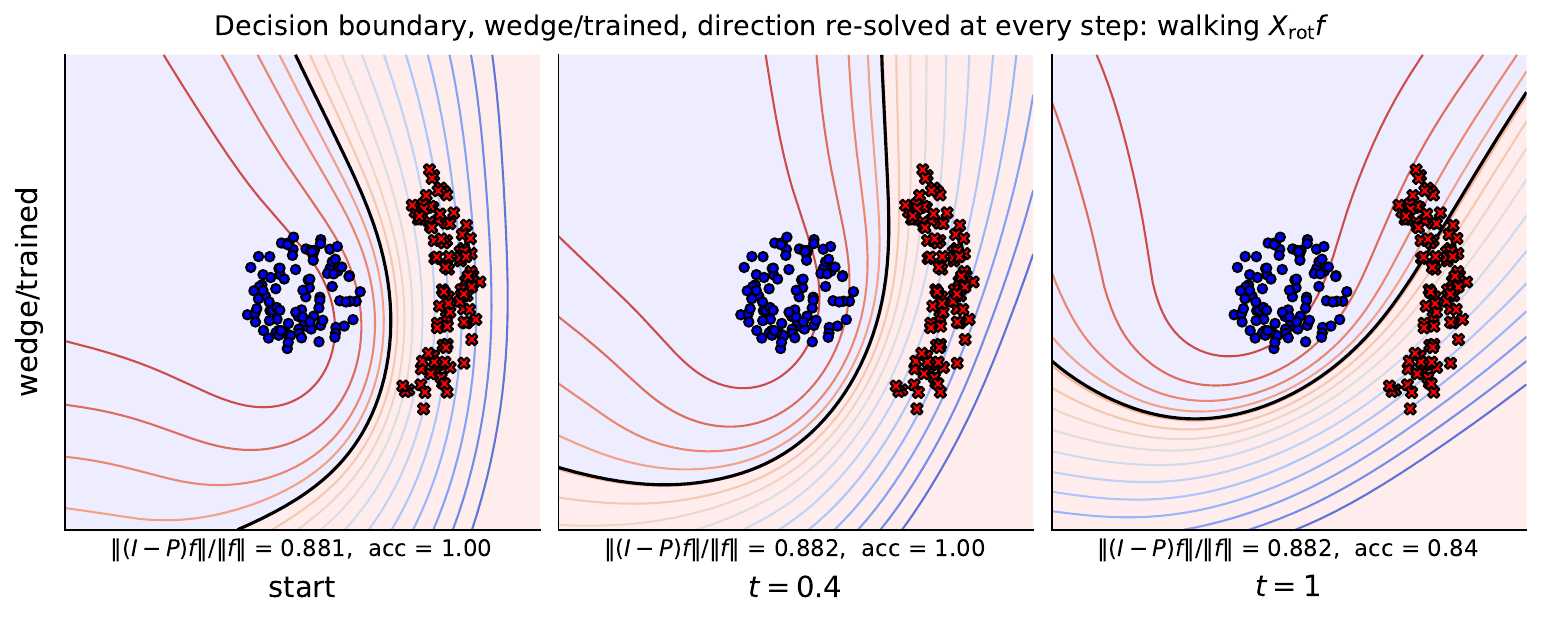}

        \smallskip
        {\small (b) Symmetry orbit flow}
    \end{minipage}

    \caption{
    Resolved parameter-space flows for the wedge-trained classifier. (a) The equivariance-directed flow along $b^\star$ moves the realised function towards the equivariant subspace, reducing its non-invariant component.
    (b) The symmetry-directed flow along $c^\star$ moves the realised function along its rotational symmetry orbit while approximately preserving its distance from the equivariant subspace.
    The local direction is recomputed at every integration step.
    }
    \label{fig:annulus_resolved}
\end{figure}

We train an MLP to distinguish an inner disc from an outer annulus, either using a $90^\circ$ wedge or the full $360^\circ$ annulus. For each model, we compute the symmetry-directed direction $c^\star$ and the equivariance-directed direction $b^\star$. Since these directions are defined locally, we compare a \textsc{Resolved} trajectory, which recomputes the Jacobian and least-squares direction at every step, with a \textsc{Fixed} trajectory that reuses the direction computed at $t=0$. Figures~\ref{fig:annulus_resolved} and~\ref{fig:annulus_fixed} show the resulting flows, while Figure~\ref{fig:angular_profiles} compares their angular deviation profiles.

The resolved trajectories closely follow the predicted function-space dynamics: the symmetry-directed flow approximately preserves the equivariance defect, whereas the equivariance directed flow reduces it. Fixed directions can deviate substantially after training, while the discrepancy is much smaller at initialisation. The construction is thus reliable locally, but must be recomputed over finite trajectories; see Appendix~\ref{app:Annulus_Additional_Results} for further results.

\subsection{Physics-Informed Neural Networks}
\label{sec:experiments_pinn}

As a dynamical benchmark, we consider the \textit{Mexican hat} potential defined by a isotropic quartic Hamiltonian
\[
    H(q,p)
    =
    \frac12\|p\|^2
    +
    \frac12\alpha\|q\|^2
    +
    \frac14\|q\|^4,
    \qquad q,p\in\mathbb R^2,
\]
with equations of motion
\(\dot q=p\) and
\(\dot p=-(\alpha+\|q\|^2)q\).

The dynamics are exactly equivariant under simultaneous rotations
\((q,p)\mapsto(R_\varphi q,R_\varphi p)\); the radial force has vanishing rotational Lie derivative, as verified in Appendix~\ref{app:MexicanHat_Additional_Results}. We use Adaptable symplectic recurrent neural networks (ASRNNs) introduced by \cite{thapar2026machinelearninghamiltoniandynamical} as a Hamiltonian neural network architecture (see Appendix \ref{asrnnarchitecture} for details) designed to learn and predict the dynamics using only trajectory data and at multiple values of $\alpha$. As before we compare a \textsc{Resolved} and \textsc{Fixed} trajectory; the angular deviation profiles at multiple values of $\alpha$ are shown in Figures~\ref{fig:mexican_hat_resolved} and~\ref{fig:mexican_hat_fixed} of Appendix~\ref{app:MexicanHat_Additional_Results}, deferred there for space. We can see clearly that when the direction is recomputed at every step, walking along $c^\star$ and $b^\star$ shifts and reduces the equivariance error respectively. The former corresponds to effectively rotating the potential, while the latter actually makes the learned potential `more' equivariant. In contrast when the direction is kept fixed, while we see similar behaviour for the first few steps, the drift accumulates over time.


\ifanonymous\else
\acks{AM is supported by a Scientific Computing PhD studentship from the Ada Lovelace Centre and expresses gratitude to the Mathematical Institute, University of Oxford for funding. VT expresses gratitude to the Rhodes Trust and the Mathematical Institute, University of Oxford for funding. TB acknowledges funding from the Klaus Tschira Foundation.}
\fi

\bibliography{references}

\appendix

\section[Symmetry Orbits and Infinitesimal Lifting]
{Symmetry Orbits and Infinitesimal Lifting}
\label{app:symmetry-realisability}

We study whether the action of a Lie group on a function space can be lifted,
at least infinitesimally, through the parameterisation of a neural network.
The group- and Lie-algebraic constructions below are standard; the object of
interest is the resulting differential lifting problem for the neural-network
realisation map.

\subsection{Motivation and Setup}

\begin{figure}[ht]
    \centering
    \includegraphics[width=0.82\linewidth]{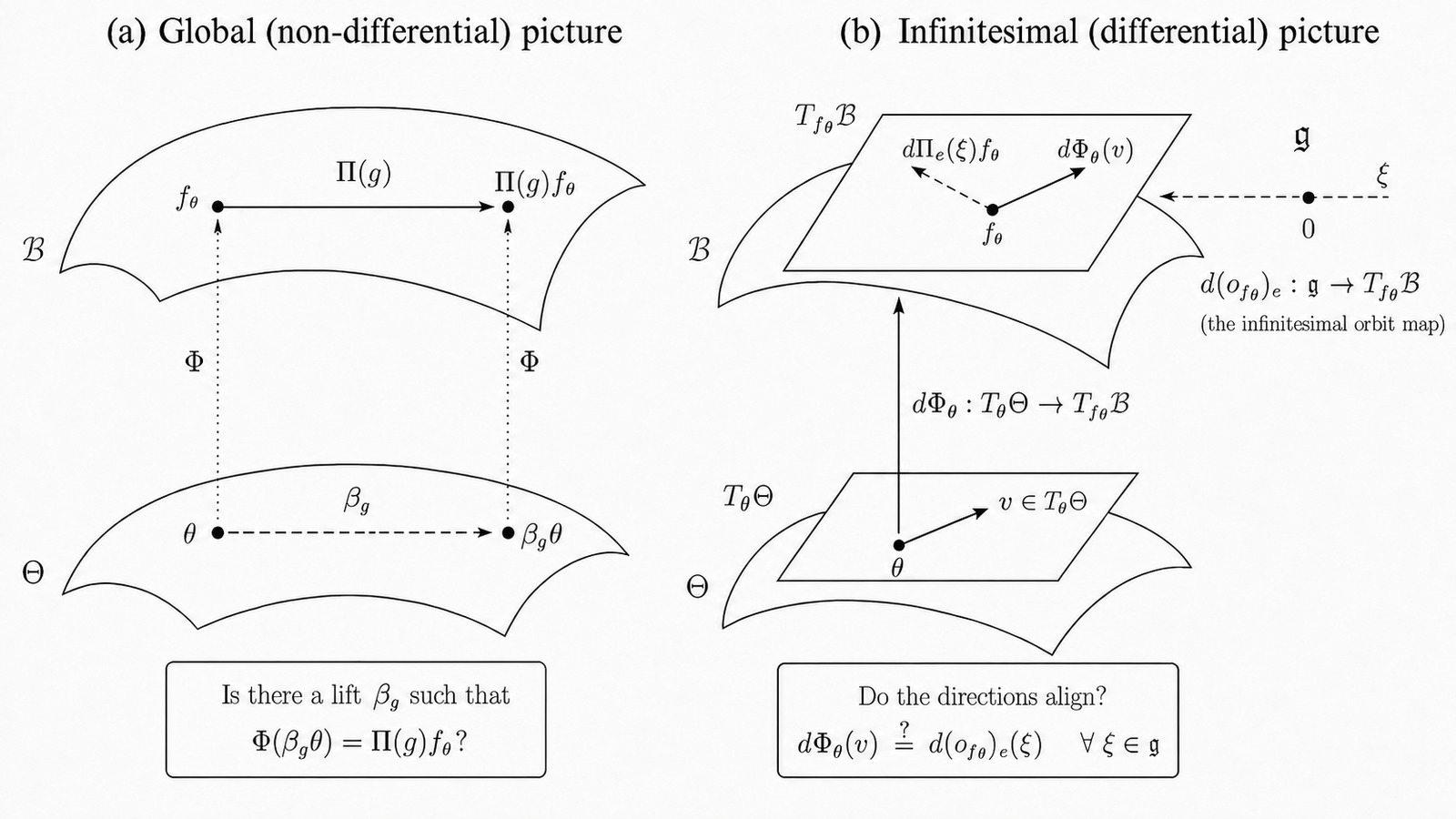}
    \caption{Global and infinitesimal lifting through the neural-network realisation map. Left: can a symmetry transformation in function space be realised by a transformation in parameter space? Right: do the infinitesimal symmetry directions lie in the function-space tangent directions accessible through the parametrisation?}
    \label{fig:conceptual-overview}
\end{figure}

Let \(G\) be a finite-dimensional Lie group acting smoothly on a manifold
\(\mathcal X\) from the left:
\[
    \tau:G\times\mathcal X\longrightarrow\mathcal X,
    \qquad
    (g,x)\longmapsto \tau_gx.
\]
Let \((\rho,V)\) be a finite-dimensional representation of \(G\), and let
\(\mathcal B\) be a \(G\)-invariant vector space of differentiable maps
\(f:\mathcal X\to V\). In function space, the actions on the input and output spaces induce a
representation on \(\mathcal B\):
\begin{equation}
    \Pi:G\longrightarrow\operatorname{GL}(\mathcal B),
    \qquad
    [\Pi(g)f](x):=\rho(g)f(\tau_{g^{-1}}x).
    \label{eq:induced-representation}
\end{equation}

A function is fixed by \(\Pi\) precisely when it is equivariant with respect
to the actions on \(\mathcal X\) and \(V\):
\begin{equation}
    f\in\mathcal B^G
    \quad\Longleftrightarrow\quad
    f(\tau_gx)=\rho(g)f(x)
    \quad
    \text{for all }g\in G,\ x\in\mathcal X.
    \label{eq:fixed-points-equivariance}
\end{equation}

This construction describes symmetry directly in function space. In a neural
network, however, functions are represented through a finite-dimensional
parameter space. We therefore introduce a smooth parameter manifold
\(\Theta\) and a realisation map
\begin{equation}
    \Phi:\Theta\longrightarrow\mathcal B,
    \qquad
    \theta\longmapsto f_\theta.
    \label{eq:realisation-map}
\end{equation}

The strongest form of the lifting problem asks whether there exists a smooth
action
\[
    \beta:G\times\Theta\longrightarrow\Theta
\]
such that
\[
    \Phi(\beta_g\theta)=\Pi(g)\Phi(\theta)
\]
for every \(g\in G\) and \(\theta\in\Theta\). In other words, we ask whether
the realisation map \(\Phi\) can be made \(G\)-equivariant.

Let
\[
    \mathcal M:=\Phi(\Theta)
\]
denote the realised model class. Although \(\Pi\) defines an action on the
ambient function space \(\mathcal B\), this action need not preserve
\(\mathcal M\). In particular, the transformed function
\(\Pi(g)f_\theta\) need not be realised by any parameter in \(\Theta\).
Even when \(\mathcal M\) is \(G\)-invariant, it may not be possible to choose
parameter representatives in a way that defines a smooth action on
\(\Theta\).

Our primary interest is the weaker infinitesimal version of this question.
At a fixed parameter value \(\theta\), can every first-order variation of
\(f_\theta\) generated by the group action be realised by a first-order
variation of the parameters? The following subsections formulate this
question by comparing the tangent space to the symmetry orbit of
\(f_\theta\) with the image of the differential of \(\Phi\).



\subsection{The Differential Lifting Criterion}
\label{sec:infinitesimal-lifting-appendix}

We now derive a necessary criterion for there to be such a lifting by comparing the infinitesimal symmetry directions in function space with
the function variations generated by changes in the network parameters. At
\(\theta\in\Theta\), the differential of the realisation map is
\begin{equation}
    J_\theta
    :=
    \mathrm d\Phi_\theta:
    T_\theta\Theta
    \longrightarrow
    T_{f_\theta}\mathcal B
    \cong \mathcal B.
    \label{eq:realisation-differential}
\end{equation}
Thus \(J_\theta v\) is the first-order change in the realised function
produced by the parameter velocity \(v\in T_\theta\Theta\).

When \(\Theta\subseteq\mathbb R^p\), this differential takes the form
\begin{equation}
    [J_\theta v](x)
    =
    \sum_{i=1}^p
    v_i S^*_i(x;\theta),
    \qquad
    S^*_i(x;\theta)
    :=
    \frac{\partial f_\theta(x)}{\partial\theta_i}.
    \label{eq:functional-sensitivities}
\end{equation}
The space of first-order function variations available through the
parameterisation is therefore
\begin{equation}
    \mathcal T_\theta
    :=
    \operatorname{Im}(J_\theta)
    =
    \operatorname{span}
    \{S^*_1(\,\cdot\,;\theta),\ldots,S^*_p(\,\cdot\,;\theta)\}.
    \label{eq:model-tangent}
\end{equation}
At a regular point of \(\Phi\), this is the tangent space
\(T_{f_\theta}\mathcal M\) of the realised model class.

Suppose now that a smooth parameter action
\[
    \beta:G\times\Theta\longrightarrow\Theta
\]
realises the function-space action. For fixed \(\theta\), define the
parameter-space orbit map
\[
    b_\theta:G\longrightarrow\Theta,
    \qquad
    b_\theta(g):=\beta_g\theta.
\]
The equivariance condition
\[
    \Phi(\beta_g\theta)=\Pi(g)\Phi(\theta)
\]
can then be written as the identity
\begin{equation}
    \Phi\circ b_\theta=o_{f_\theta},
    \label{eq:orbit-map-identity}
\end{equation}
where \(o_{f_\theta}(g)=\Pi(g)f_\theta\) is the function-space orbit map.

Differentiating \eqref{eq:orbit-map-identity} at the identity
\(e\in G\) and applying the chain rule gives
\begin{equation}
    \mathrm d\Phi_\theta
    \circ
    \mathrm d(b_\theta)_e
    =
    \mathrm d(o_{f_\theta})_e.
    \label{eq:differentiated-orbit-map}
\end{equation}
Define
\[
    L_\theta
    :=
    \mathrm d(b_\theta)_e:
    \mathfrak g\longrightarrow T_\theta\Theta.
\]
For \(\xi\in\mathfrak g\), \(L_\theta(\xi)\) is the parameter velocity
generated by \(\xi\). Using the definitions
\(J_\theta=\mathrm d\Phi_\theta\) and
\(a_{f_\theta}=\mathrm d(o_{f_\theta})_e\), equation
\eqref{eq:differentiated-orbit-map} becomes
\begin{equation}
    J_\theta L_\theta
    =
    a_{f_\theta}.
    \label{eq:linear-lift}
\end{equation}
Equivalently, for every \(\xi\in\mathfrak g\),
\begin{equation}
    J_\theta L_\theta(\xi)
    =
    a_{f_\theta}(\xi)
    =
    \mathrm d\Pi(\xi)f_\theta.
    \label{eq:infinitesimal-equivariance}
\end{equation}

Equation \eqref{eq:linear-lift} immediately yields the necessary condition
\begin{equation}
    \operatorname{Im}(a_{f_\theta})
    \subseteq
    \operatorname{Im}(J_\theta).
    \label{eq:infinitesimal-lifting-criterion}
\end{equation}
Using
\[
    \operatorname{Im}(a_{f_\theta})
    =
    T_{f_\theta}(G\!\cdot\!f_\theta),
    \qquad
    \operatorname{Im}(J_\theta)
    =
    \mathcal T_\theta,
\]
this condition can equivalently be written as
\begin{equation}
    T_{f_\theta}(G\!\cdot\!f_\theta)
    \subseteq
    \mathcal T_\theta.
    \label{eq:orbit-tangent-inclusion-appendix}
\end{equation}

Thus, if a smooth parameter action realising \(\Pi\) exists, every
infinitesimal function-space orbit direction must be attainable through an
infinitesimal change of the parameters.

Conversely, at a fixed parameter value \(\theta\), the inclusion
\eqref{eq:infinitesimal-lifting-criterion} is sufficient for the existence of
some linear map
\[
    L_\theta:\mathfrak g\longrightarrow T_\theta\Theta
\]
satisfying \eqref{eq:linear-lift}. It is therefore necessary and sufficient
for pointwise first-order lifting. It is not sufficient for the existence of
a local or global action on \(\Theta\): such an action requires the lifts to
vary smoothly with \(\theta\), satisfy the appropriate Lie-bracket relations,
and obey the relevant integrability conditions.

\subsection{A Least-Squares Lifting Objective}
\label{sec:approximate-lifting-appendix}

The differential lifting condition requires the operator equation
\[
    J_\theta L_\theta=a_{f_\theta}
\]
to have a solution. When this equation cannot be solved exactly, its failure
can be quantified through a least-squares relaxation.

Assume that \(\mathcal B\) is equipped with a Hilbert-space inner product,
and choose inner products on \(\mathfrak g\) and \(T_\theta\Theta\). For a
candidate linear lift
\[
    L:\mathfrak g\longrightarrow T_\theta\Theta,
\]
define the lifting error
\begin{equation}
    \mathrm{Err}_\theta(L)
    :=
    \left\|
        J_\theta L-a_{f_\theta}
    \right\|_{\mathrm{HS}}^2.
    \label{eq:lifting-least-squares-error}
\end{equation}
If \(\{\xi_a\}_{a=1}^d\) is an orthonormal basis of \(\mathfrak g\), then
\begin{equation}
    \mathrm{Err}_\theta(L)
    =
    \sum_{a=1}^d
    \left\|
        J_\theta L(\xi_a)-a_{f_\theta}(\xi_a)
    \right\|_{\mathcal B}^2.
    \label{eq:lifting-error-generators}
\end{equation}

Let \(P_\theta\) denote the orthogonal projector onto
\(\mathcal T_\theta=\operatorname{Im}(J_\theta)\). Since
\(J_\theta L\) takes values in \(\mathcal T_\theta\), we have the orthogonal
decomposition
\begin{equation}
    \begin{aligned}
        \mathrm{Err}_\theta(L)
        &=
        \left\|
            J_\theta L-P_\theta a_{f_\theta}
        \right\|_{\mathrm{HS}}^2
        +
        \left\|
            (I-P_\theta)a_{f_\theta}
        \right\|_{\mathrm{HS}}^2.
    \end{aligned}
    \label{eq:lifting-error-decomposition}
\end{equation}
The second term is independent of \(L\). Hence
\begin{equation}
    \min_L \mathrm{Err}_\theta(L)
    =
    \left\|
        (I-P_\theta)a_{f_\theta}
    \right\|_{\mathrm{HS}}^2 .
    \label{eq:minimum-lifting-error}
\end{equation}
The minimum lifting error is therefore exactly the part of the orbit map that
is orthogonal to the model tangent space, and it vanishes if and only if the
differential lifting criterion
\eqref{eq:infinitesimal-lifting-criterion} holds. It is a genuinely
function-space quantity: it measures what the parametrisation cannot express,
not how well any particular optimiser performs.

Because \eqref{eq:lifting-error-generators} decouples over an orthonormal
basis of \(\mathfrak g\), the minimiser can be computed one generator at a
time. Writing \(\eta_\xi:=a_{f_\theta}(\xi)\) for the function-space
direction induced by \(\xi\in\mathfrak g\) (the symbol \(g\) being reserved
for group elements), the minimum-norm minimiser is
\begin{equation}
    L^\star_\theta(\xi)
    =
    c^\star(\xi)
    :=
    J_\theta^{+}\eta_\xi ,
    \label{eq:generatorwise-minimiser}
\end{equation}
which is \eqref{eq:min_norm_attribution} of the main text. Its components
\(c^\star_i\) are the per-parameter attribution of the symmetry generator
\(\xi\), and are meaningful precisely because the columns of \(J_\theta\)
are the pointwise functional sensitivities \(S^*_i(\,\cdot\,;\theta)\) of
\eqref{eq:functional-sensitivities}.

\subsection{The Equivariance Defect and its Attribution}
\label{app:defect}

The objective above moves \emph{along} the symmetry orbit. To move
\emph{towards} equivariance we use the Reynolds projector onto the fixed-point
subspace \(\mathcal E:=\mathcal B^G\). For compact \(G\) with normalised Haar
measure \(\mathrm d\mu\),
\begin{equation}
    [P_{\mathcal E}f](x)
    :=
    \int_G \rho(g)^{-1} f(\tau_g x)\,\mathrm d\mu(g),
    \qquad
    \delta_\theta
    :=
    f_\theta-P_{\mathcal E}f_\theta ,
    \label{eq:reynolds}
\end{equation}
so that \(P_{\mathcal E}\) is the orthogonal projection onto \(\mathcal E\)
and \(\delta_\theta\) is the equivariance defect. The two objectives are
complementary rather than redundant: since \(P_{\mathcal E}\) annihilates
every \(\mathrm d\Pi(\xi)\)-direction, \(\mathcal E=\ker\mathrm d\Pi(\xi)\)
for each \(\xi\), and a short calculation using the skew-adjointness of
\(\mathrm d\Pi(\xi)\) on \(\mathcal B\) (valid for unitary \(\rho\) and
measure-preserving \(\tau\)) gives
\begin{equation}
    \langle \delta_\theta,\ \eta_\xi\rangle_{\mathcal B}
    =
    \langle \delta_\theta,\ \mathrm d\Pi(\xi)\delta_\theta\rangle_{\mathcal B}
    =
    0 .
    \label{eq:orthogonality}
\end{equation}
The orbit direction \(\eta_\xi\) is thus tangent to the level set of
\(\lVert\delta_\theta\rVert\) through \(f_\theta\): to first order it rotates
the learned function without changing how equivariant it is, whereas
\(-\delta_\theta\) is the direction of steepest decrease of the defect. This
is what the two flows of Section~\ref{sec:approximate-lifting} are predicted
to do, and what the experiments test. Relaxing the second problem in the same
least-squares sense gives \(b^\star=J_\theta^{+}(-\delta_\theta)\), i.e.
\eqref{eq:min_norm_delta}.

\subsection{Numerical Realisation}
\label{app:numerics}

All of the above is stated in an infinite-dimensional \(\mathcal B\); in
practice every function-space object is discretised on a finite probe set
\(\{x_n\}_{n=1}^{N}\subset\mathcal X\), and \(\mathcal B\) is given the
corresponding empirical \(L^2\) inner product. This makes
\(J_\theta\) an explicit matrix
\(J\in\mathbb R^{N\dim\mathcal Y\times p}\) whose \((n,a),i\) entry is
\(\partial f_{\theta,a}(x_n)/\partial\theta_i\), obtained by automatic
differentiation, and makes \(\eta_\xi\) and \(\delta_\theta\) explicit vectors
in \(\mathbb R^{N\dim\mathcal Y}\). The aggregate sensitivity of
Section~\ref{sec:Introduction} is then the column norm
\(S_i(\theta)=\lVert J_{:,i}\rVert_2/\sqrt{N\dim\mathcal Y}\), i.e. the
root-mean-square of \(\partial f_\theta/\partial\theta_i\) over probe points
and output components. Note that the expectation is taken \emph{after} the
pointwise magnitude: averaging the signed sensitivity would let a parameter
whose local effect changes sign across \(\mathcal X\) cancel to zero. \(S_i\)
is a scalar summary of the vector field \(S^*_i(\,\cdot\,;\theta)\); it is the
latter that spans the accessible tangent directions \eqref{eq:model-tangent}.

\paragraph{Truncation of \(J_\theta^{+}\).}
\(\mathcal T_\theta\) is taken to be the \emph{numerically resolved} tangent
space. Writing the singular values of \(J\) as
\(\sigma_1\ge\cdots\ge\sigma_{\min(N\dim\mathcal Y,p)}\), we retain only those
directions with
\begin{equation}
    \sigma_k \ \ge\ \tau\,\sigma_1,
    \qquad
    \tau = 10^{-3},
    \label{eq:svd-cutoff}
\end{equation}
and \(J_\theta^{+}\) denotes the pseudoinverse of the correspondingly
truncated SVD. Without truncation an overparameterised network declares every
direction reachable, since \(J\) generically has many tiny but nonzero
singular values on a finite probe grid; \(\tau\) fixes the scale below which a
direction is treated as noise. All solves are performed in float64.

\paragraph{Gauge of the attribution coefficients.}
The raw sensitivity \(S_i\) is not invariant under reparametrisation: under
\(\theta_i\mapsto\lambda\theta_i\) (for instance the positive-homogeneity
rescaling of a ReLU layer, \(W^{(\ell)}\mapsto\lambda W^{(\ell)}\),
\(W^{(\ell+1)}\mapsto\lambda^{-1}W^{(\ell+1)}\), which leaves \(f_\theta\)
unchanged) one has \(S_i\mapsto\lambda^{-1}S_i\), and the plain minimum-\(\ell_2\)-norm
solution of \eqref{eq:min_norm_attribution} would correspondingly reallocate
attribution towards parameters that merely happen to carry small columns.
We therefore minimise the norm in the sensitivity-normalised gauge: with
\(D:=\operatorname{diag}(\lVert J_{:,1}\rVert_2,\dots,\lVert J_{:,p}\rVert_2)\),
we solve
\begin{equation}
    \tilde c^\star=(JD)^{+}\,\eta_\xi,
    \qquad
    c^\star=D^{-1}\tilde c^\star ,
    \label{eq:normalised-gauge}
\end{equation}
so that the quantity actually minimised is \(\sum_i S_i^2 (c^\star_i)^2\),
which is invariant under the rescalings above. The projection
\(J c^\star\), the residual and the resolved rank are unaffected by this
choice --- only the split of a fixed function-space direction across
parameters is. Columns with
\(\lVert J_{:,i}\rVert_2\le 10^{-8}\max_j\lVert J_{:,j}\rVert_2\) are treated
as exactly dead and assigned zero attribution, rather than having
machine-precision noise amplified by \(D^{-1}\).

\paragraph{Probe sets and quadrature.}
For the Hamiltonian experiments, \(f_\theta\) is the learned force field
\(F_\theta(q;\alpha)=-\nabla_q V_{\theta_2}(q;\alpha)\), probed on a
rotationally symmetric polar grid of \(15\) radii \(\times\ 32\) angles
(\(N=480\) points) in \(\lVert q\rVert\le 1\); a square grid is not invariant
under the group and was found to leak cross-talk between \(\eta_\xi\) and
\(\delta_\theta\). The Reynolds average \eqref{eq:reynolds} is evaluated by
the \(64\)-point equally spaced quadrature rule on \(SO(2)\), which is exact
up to the truncation of the Fourier series and reproduces an exactly
equivariant field to \(\sim\!10^{-15}\) relative error. Rows for all training
and analysis values of \(\alpha\) are stacked into a single least-squares
system, so that one direction is obtained per step rather than an
uncoordinated direction per \(\alpha\). Finite-rotation equivariance errors
are reported at a fixed angle of \(73^\circ\).

\section{Supplementary Material}
\label{sec:Supplementary_Material}

\subsection{ASRNN architecture} \label{asrnnarchitecture} 

Learning Hamiltonian systems using neural networks has received considerable attention in recent years. Hamiltonian neural networks (HNNs), an early architecture for this problem, were introduced by Greydanus \textit{et al.}~\cite{greydanus2019hamiltonian}. These models parametrise a time-independent Hamiltonian $\mathcal{H} \approx \mathcal{H}_{\theta}(q,p)$ using an MLP with parameters $\theta$. The model is then trained using the loss 
\begin{equation}
    \mathcal{L}_{(\mathbf{q}, \mathbf{p})} (\theta) = \norm{\frac{\partial \mathcal{H}_{\theta}}{\partial \mathbf{p}} - \frac{\partial \mathbf{q}}{\partial t}}_2 + \norm{\frac{\partial \mathcal{H}_{\theta}}{\partial \mathbf{q}} + \frac{\partial \mathbf{p}}{\partial t}}_2,
\end{equation}
which enforces Hamilton's equations. As such models require accurate time derivatives in the training data, Chen \textit{et al.}~\cite{Chen2020Symplectic} introduced symplectic recurrent neural networks (SRNNs), which use symplectic integrators to predict trajectories from time derivatives obtained from the HNN gradients. This allows training with a loss that depends only on the observed trajectory. Such integrators are \textit{structure-preserving}: they preserve the symplectic structure, and consequently phase-space volume, and typically exhibit bounded long-time energy error. To enable learning across varying physical parameters, Thapar \textit{et al.}~\cite{thapar2026machinelearninghamiltoniandynamical} introduced adaptable SRNNs (ASRNNs), which incorporate parameter channels into SRNNs and have been shown to learn effectively across parameter space from sparse data.

In this work, we use ASRNNs with a leapfrog integrator~\cite{Leimkuhler_Reich_2005} as our illustrative architecture. ASRNNs parametrise a time-independent separable Hamiltonian using separate neural networks for the kinetic and potential energies, respectively. Specifically, the Hamiltonian is parametrised as $\mathcal{H}(\mathbf{q}, \mathbf{p} ; \alpha) \approx \mathcal{K}_{\theta_1}(\mathbf{p}) + \mathcal{V}_{\theta_2} (\mathbf{q}; \alpha)$, where each network is a simple MLP. Recurrence is introduced through a leapfrog integrator. Given $(\mathbf{q}_t, \mathbf{p}_t)$, the next time step is predicted as
\begin{gather}
    \mathbf{p}\left(t+\frac{\Delta t}{2}\right) = \mathbf{p}(t) - \frac{\Delta t}{2} \frac{\partial \mathcal{V}_{\theta_2}}{\partial \mathbf{q}} \bigg|_t \\
    \mathbf{q}\left( t+\Delta t\right) = \mathbf{q}(t) + \Delta t \frac{\partial \mathcal{K}_{\theta_1}}{\partial \mathbf{p}} \bigg|_{t+\frac{\Delta t}{2}} \\
    \mathbf{p}\left( t+\Delta t\right) =  \mathbf{p}\left( t+\frac{\Delta t}{2}\right) - \frac{\Delta t}{2} \frac{\partial \mathcal{V}_{\theta_2}}{\partial \mathbf{q}} \bigg|_{t+\Delta t},
\end{gather}
where $\Delta t$ is a chosen time step. Training minimizes the trajectory-matching loss,
\begin{equation}
    \mathcal{L}_{(\mathbf{q}_0, \mathbf{p}_0; \alpha)} (\theta) = \sum_{t \in \mathcal{T}} \left(\norm{\mathbf{q}(t) - \mathbf{\hat{q}(t)}}_2 + \norm{\mathbf{p}(t) - \mathbf{\hat{p}(t)}}_2  \right),
    \label{ASRNNloss}
\end{equation}
where $(\mathbf q_0,\mathbf p_0;\alpha)$ specifies the initial conditions and physical parameters, $(\mathbf q(t),\mathbf p(t))$ and $(\hat{\mathbf q}(t),\hat{\mathbf p}(t))$ denote the model predictions and ground-truth values at time $t$, respectively, and $\mathcal T$ is the set of observed time points. In our experiments, the potential and kinetic energy networks were implemented as MLPs with three hidden layers of width 32, following the hyperparameter regime used by \cite{thapar2026machinelearninghamiltoniandynamical}.

\section{Additional Results}

\subsection{Mexican Hat}
\label{app:MexicanHat_Additional_Results}

The Mexican hat dynamics of Section~\ref{sec:experiments_pinn} are exactly
equivariant under simultaneous rotations of $q$ and $p$. For the infinitesimal
rotation generator $\Omega=\left(\begin{smallmatrix}0&-1\\1&0\end{smallmatrix}\right)$,
the radial force \(F(q)=-(\alpha+\|q\|^2)q\) satisfies
$\frac{\partial F}{\partial q}(\Omega q)-\Omega F(q)=0$, i.e. it has vanishing
rotational Lie derivative, so the symmetry is a property of the true system and
not an artefact of the architecture.

\begin{figure}[h]
    \centering
    \includegraphics[width=\linewidth]{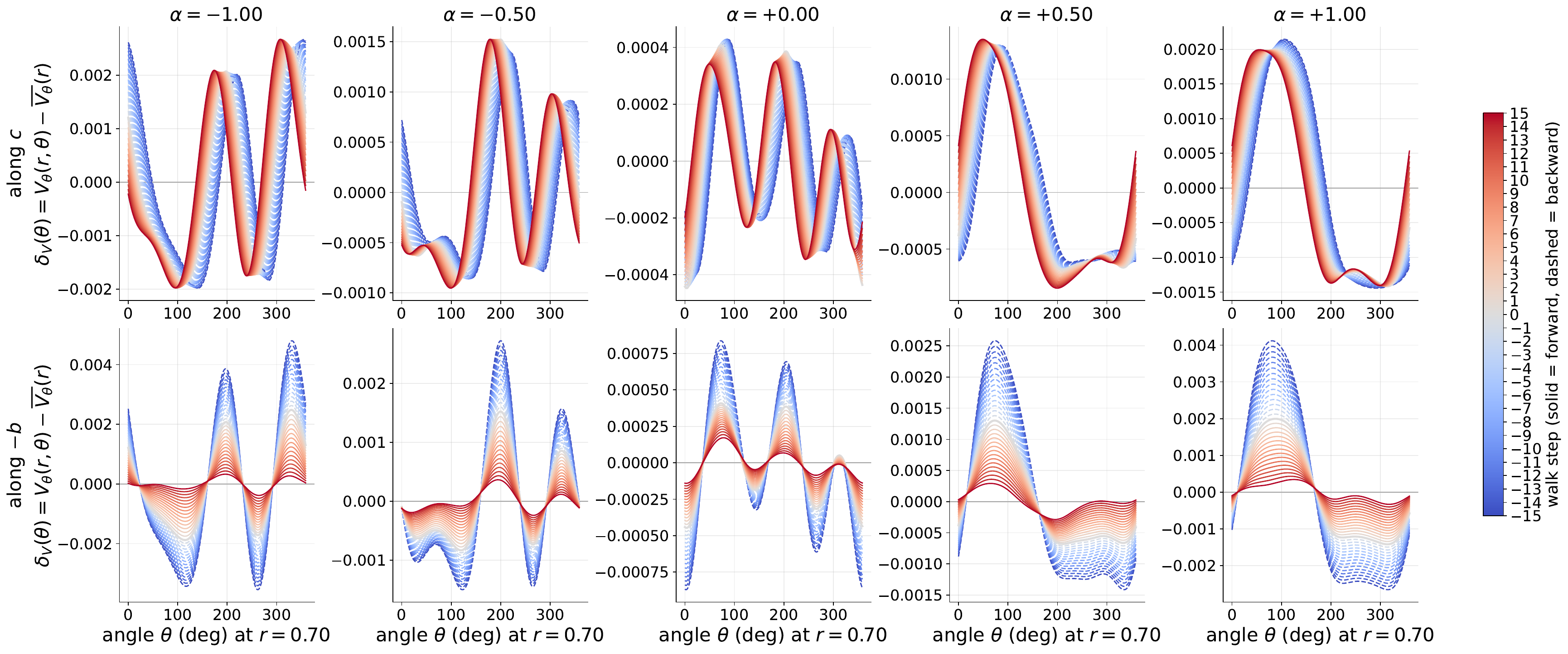}
    \caption{Angular deviation profiles for the ASRNN trained to model the Mexican hat potential for various values of $\alpha$ unseen during training. 
    Top: symmetry-directed flow along $c^\star$.
    Bottom: equivariance-directed flow along $b^\star$.
    Here the local direction is recomputed at every step, i.e. the \textsc{Resolved} trajectory.}
    \label{fig:mexican_hat_resolved}
\end{figure}

\begin{figure}[h]
    \centering
    \includegraphics[width=\linewidth]{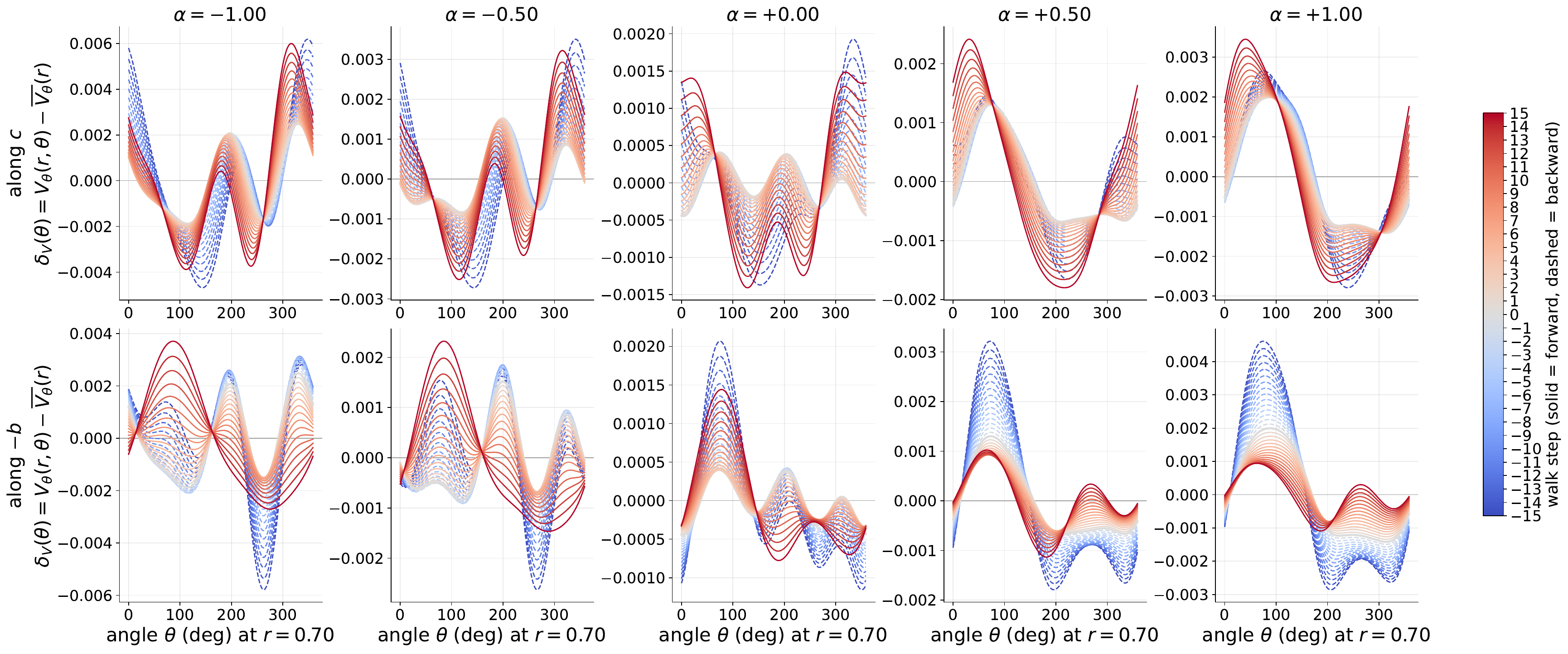}
    \caption{Angular deviation profiles for the ASRNN trained to model the Mexican hat potential for various values of $\alpha$ unseen during training. 
    Top: symmetry-directed flow along $c^\star$.
    Bottom: equivariance-directed flow along $b^\star$.
    Here the direction is computed once at $t=0$ and kept fixed, i.e. the \textsc{Fixed} trajectory.}
    \label{fig:mexican_hat_fixed}
\end{figure}

Figures~\ref{fig:mexican_hat_resolved} and~\ref{fig:mexican_hat_fixed} give the
angular deviation profiles for the ASRNN discussed in
Section~\ref{sec:experiments_pinn}, for the \textsc{Resolved} and \textsc{Fixed}
trajectories respectively.

\subsection{Annulus}
\label{app:Annulus_Additional_Results}

\begin{figure}[t]
    \centering

    \begin{minipage}[t]{0.75\linewidth}
        \centering
        \includegraphics[width=0.62\linewidth]
        {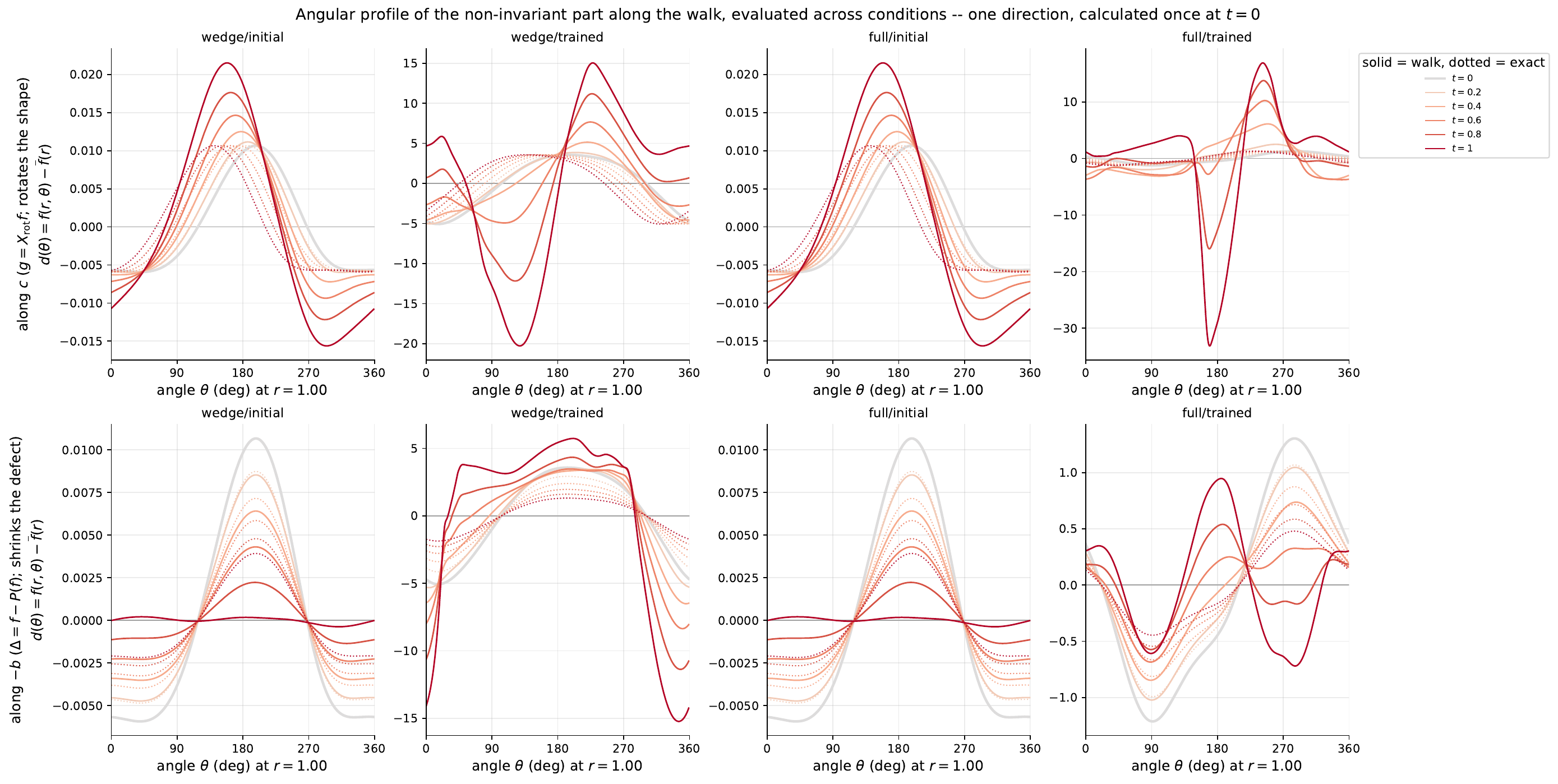}

        \smallskip
        {\small (a) Angular deviation profiles}
    \end{minipage}

    \medskip

    \begin{minipage}[t]{0.75\linewidth}
        \centering
        \includegraphics[width=0.62\linewidth]
        {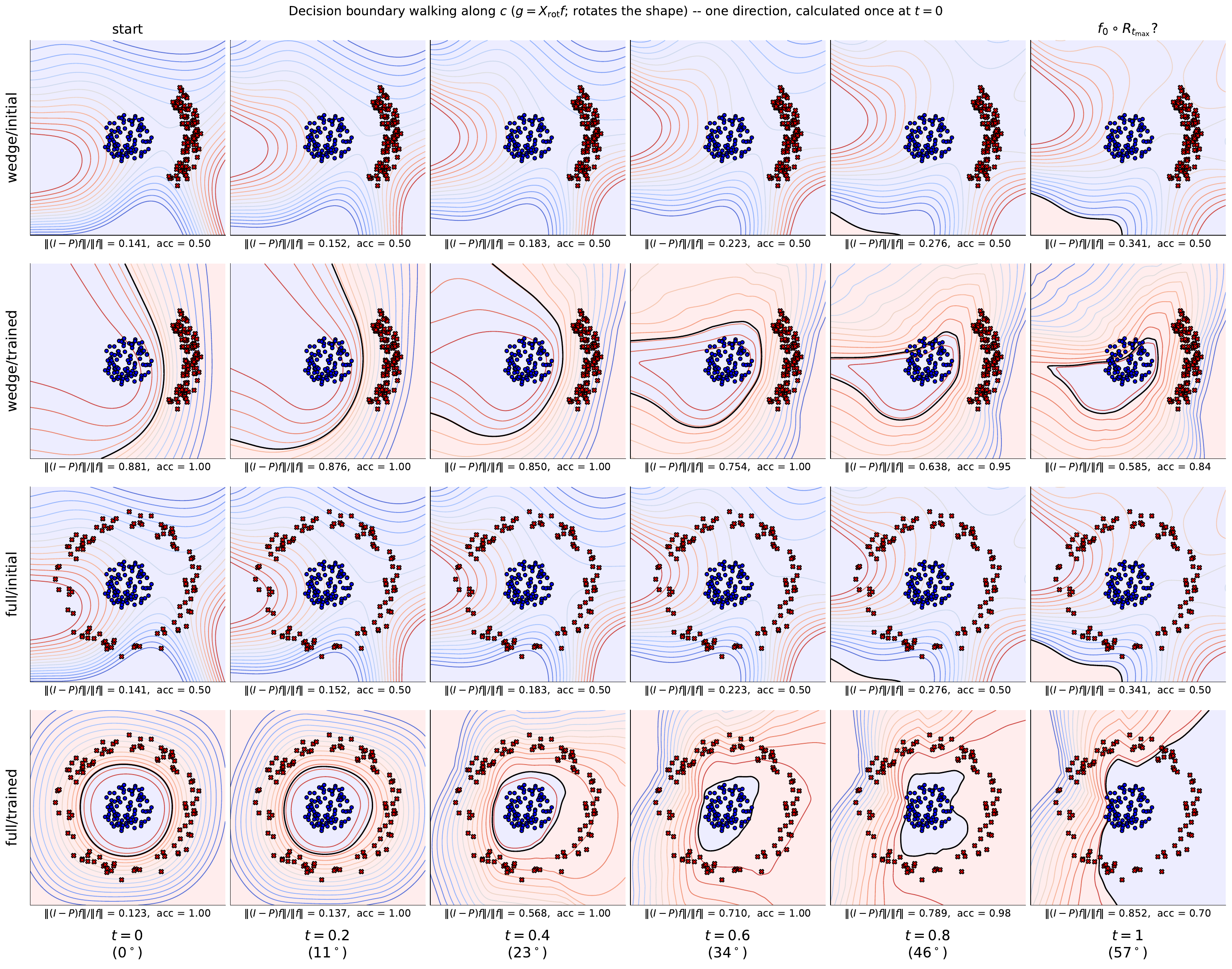}

        \smallskip
        {\small (b) Symmetry-directed flow}
    \end{minipage}

    \medskip

    \begin{minipage}[t]{0.75\linewidth}
        \centering
        \includegraphics[width=0.62\linewidth]
        {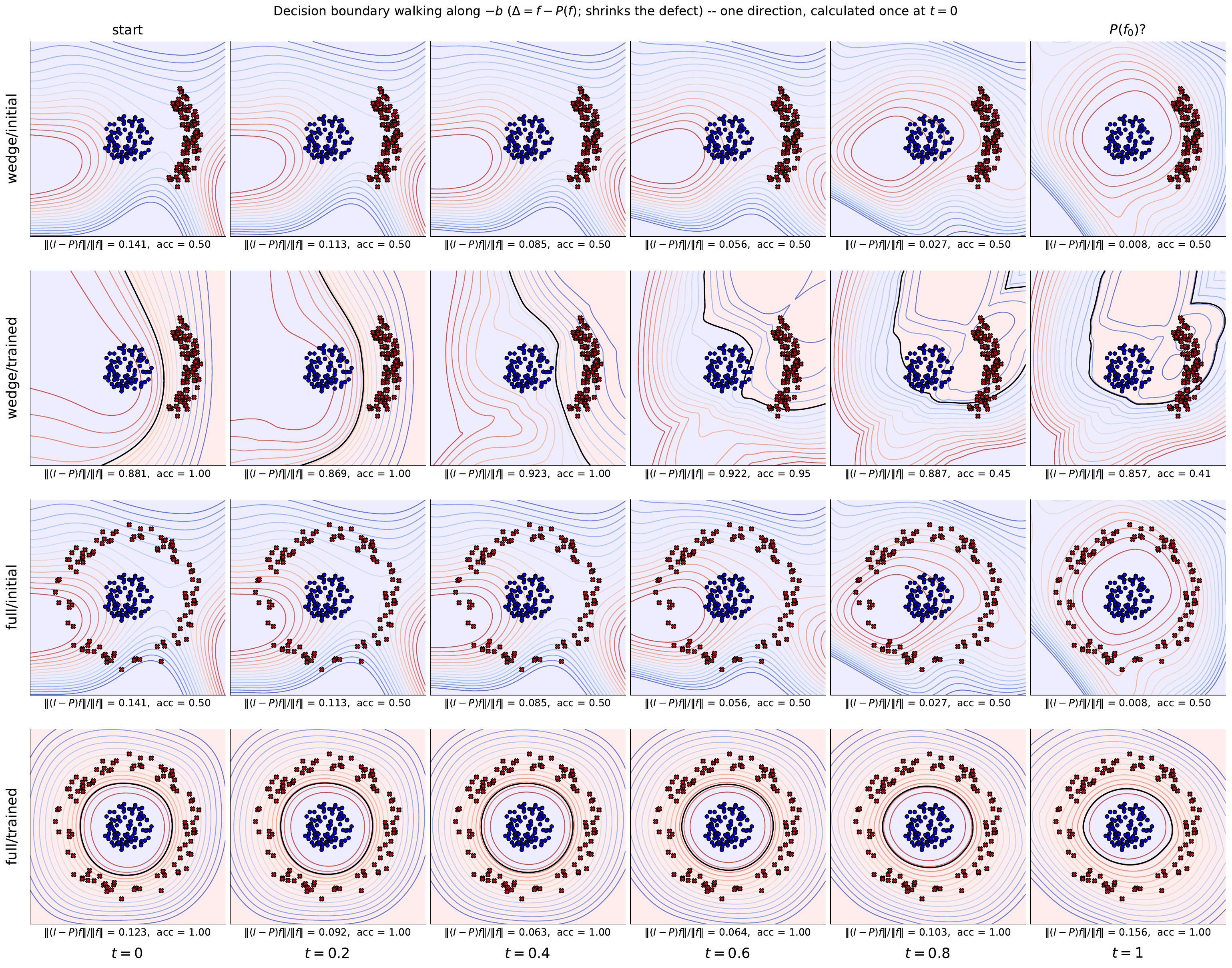}

        \smallskip
        {\small (c) Invariance-directed flow.}
    \end{minipage}

    \caption{
    Fixed parameter-space flows across training conditions.
    (a) Angular deviation profiles along the symmetry- and equivariance-directed flows.
    (b) Decision boundaries along the symmetry-directed flow $c^\star$.
    (c) Decision boundaries along the equivariance-directed flow $b^\star$.
    The parameter-space directions are computed once at $t=0$ and kept fixed throughout the trajectory.
    }
\label{fig:annulus_fixed_all}
\end{figure}

\begin{figure}[t]
    \centering

    \begin{minipage}[t]{0.75\linewidth}
        \centering
        \includegraphics[width=0.62\linewidth]
        {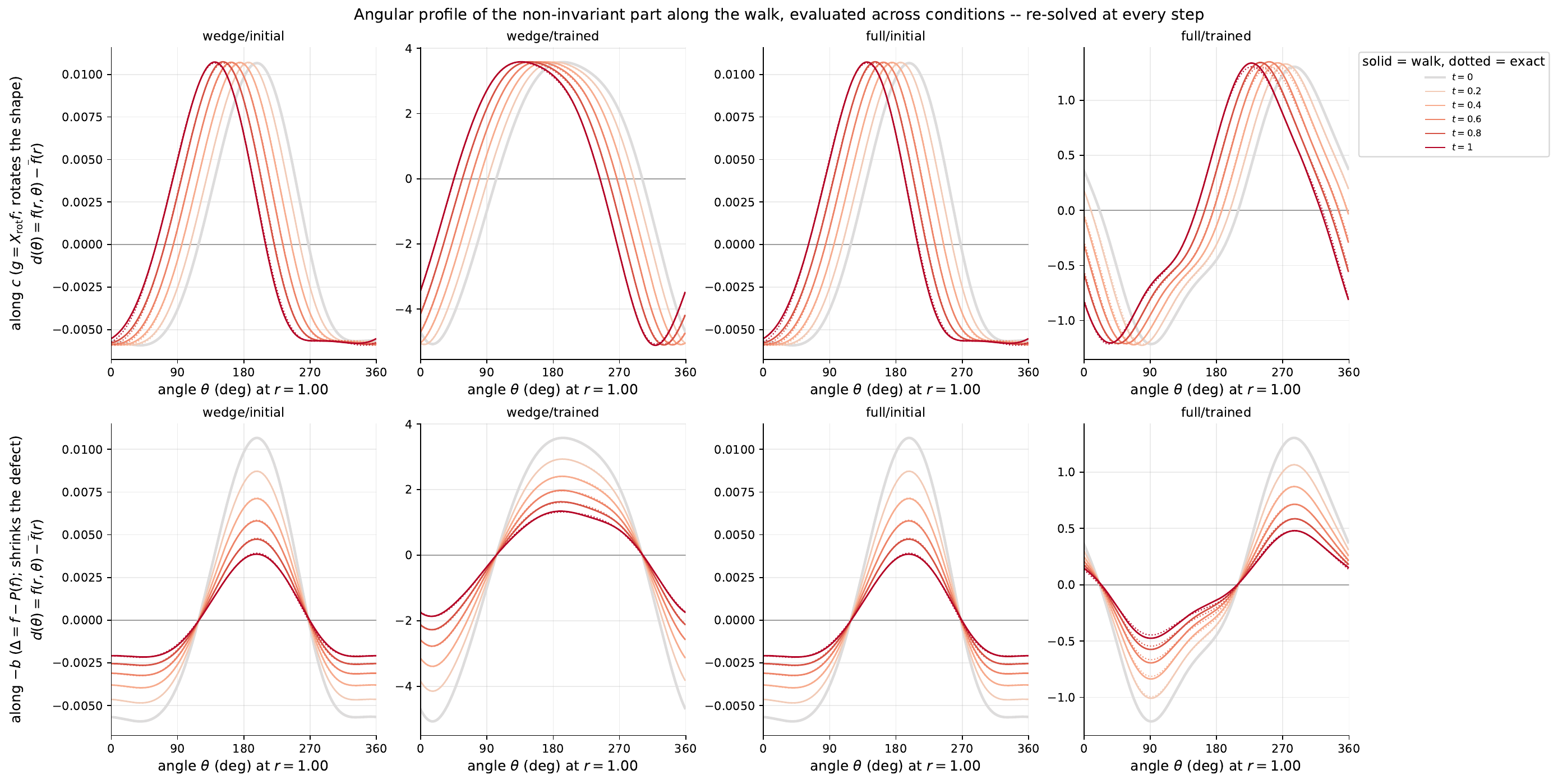}

        \smallskip
        {\small (a) Angular deviation profiles}
    \end{minipage}

    \medskip

    \begin{minipage}[t]{0.75\linewidth}
        \centering
        \includegraphics[width=0.62\linewidth]
        {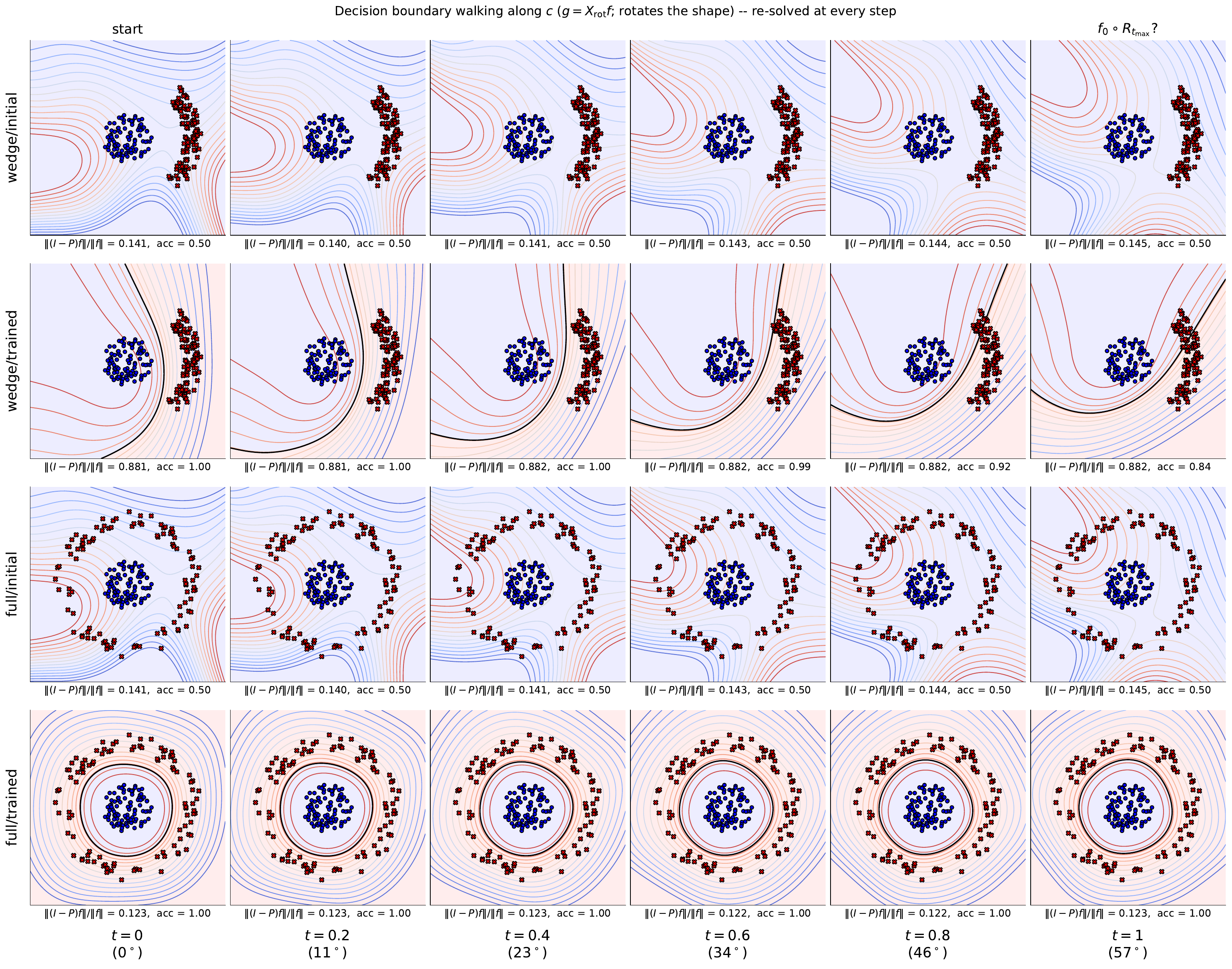}

        \smallskip
        {\small (b) Symmetry-directed flow}
    \end{minipage}

    \medskip

    \begin{minipage}[t]{0.75\linewidth}
        \centering
        \includegraphics[width=0.62\linewidth]
        {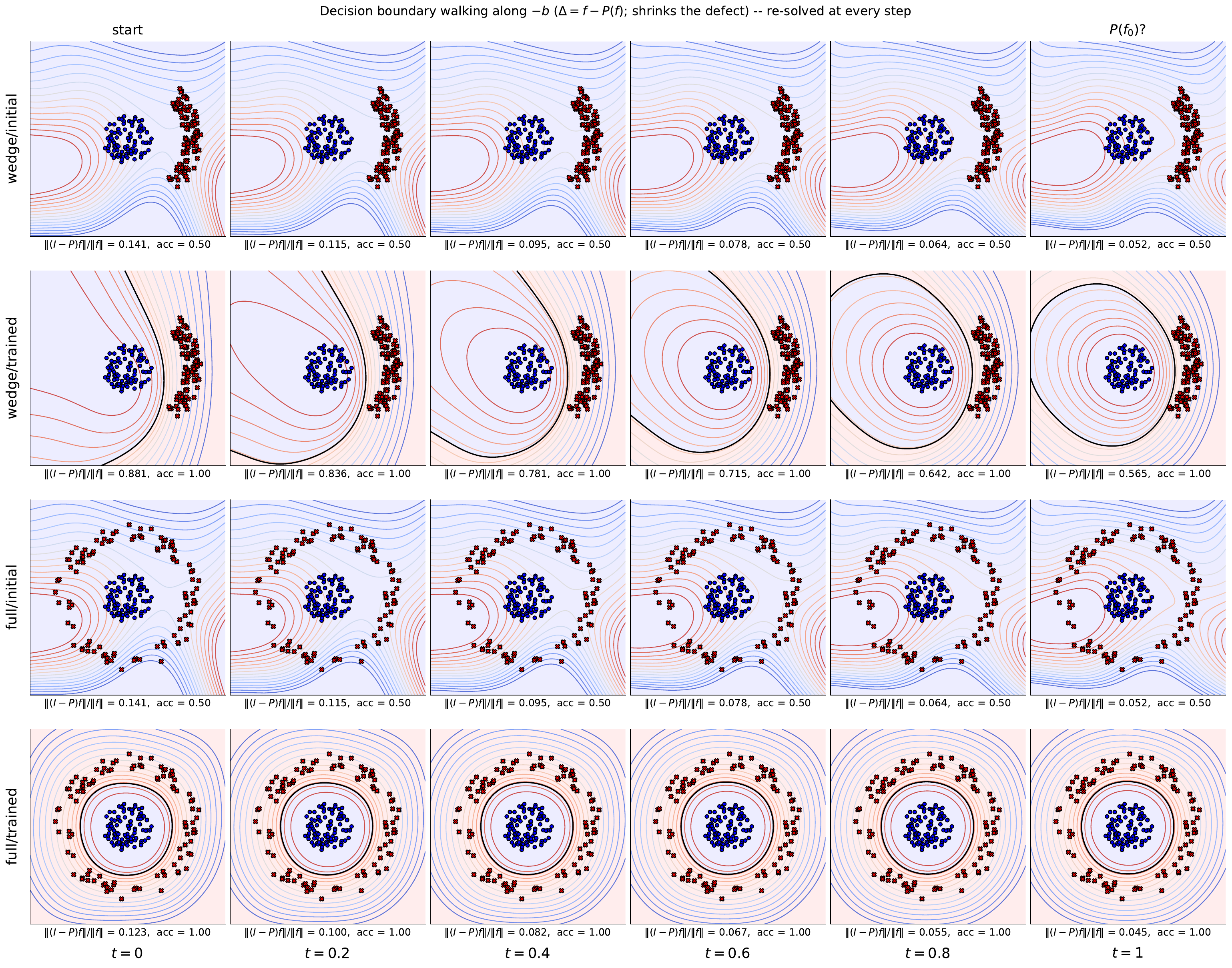}

        \smallskip
        {\small (c) Invariance-directed flow}
    \end{minipage}

    \caption{
    Resolved parameter-space flows across training conditions.
    (a) Angular deviation profiles along the symmetry- and equivariance-directed flows.
    (b) Decision boundaries along the symmetry-directed flow $c^\star$.
    (c) Decision boundaries along the equivariance-directed flow $b^\star$.
    The Jacobian and corresponding parameter-space directions are recomputed at every step.
    }
    \label{fig:annulus_resolved_all}
\end{figure}

We provide additional annulus results across initialisation and trained models,
using either the $90^\circ$ wedge or full $360^\circ$ training data.
Figures~\ref{fig:annulus_fixed_all} and~\ref{fig:annulus_resolved_all} compare
fixed and resolved parameter-space flows for the symmetry-directed direction
$c^\star$ and equivariance-directed direction $b^\star$.

At initialisation, fixed and resolved trajectories behave similarly. After
training, fixed directions can deviate substantially from their intended
function-space evolution, whereas recomputing the local direction at each step
more faithfully follows the symmetry orbit or reduces the equivariance defect.
This further illustrates that the lifting directions are inherently local.

\begin{figure}[t]
    \centering
    \includegraphics[width=0.38\linewidth]
    {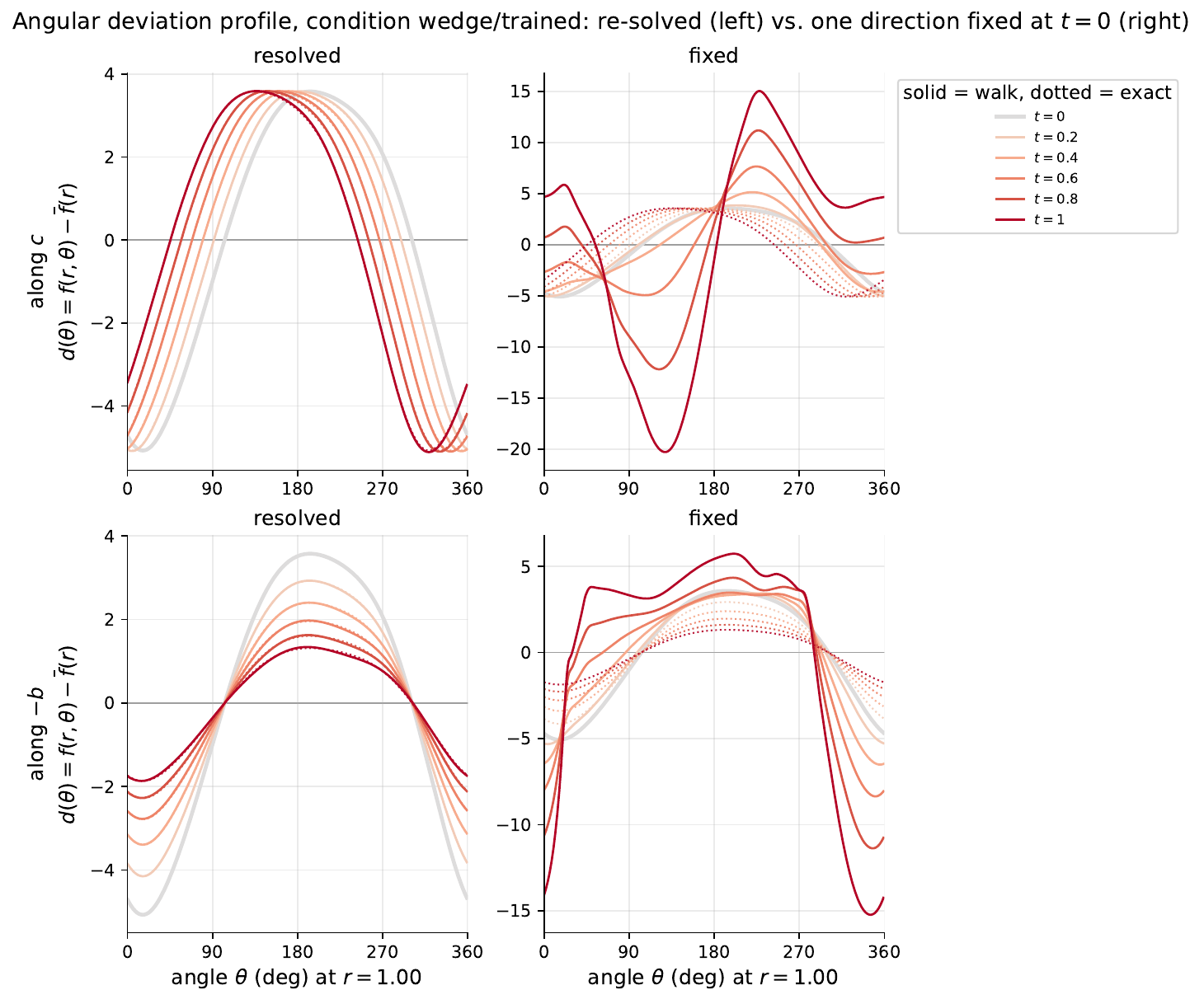}
    \caption{Angular deviation profiles for the wedge-trained classifier. Top: symmetry-directed flow along $c^\star$; bottom: equivariance-directed flow along $b^\star$. Left: the local direction is recomputed at every step; right: the initial direction is kept fixed. Solid curves denote the realised trajectory, dotted curves the ideal function-space evolution.}
    \label{fig:angular_profiles}
\end{figure}

\begin{figure}[t]
    \centering
    \begin{minipage}[t]{0.48\linewidth}
        \centering
        \includegraphics[width=\linewidth]
        {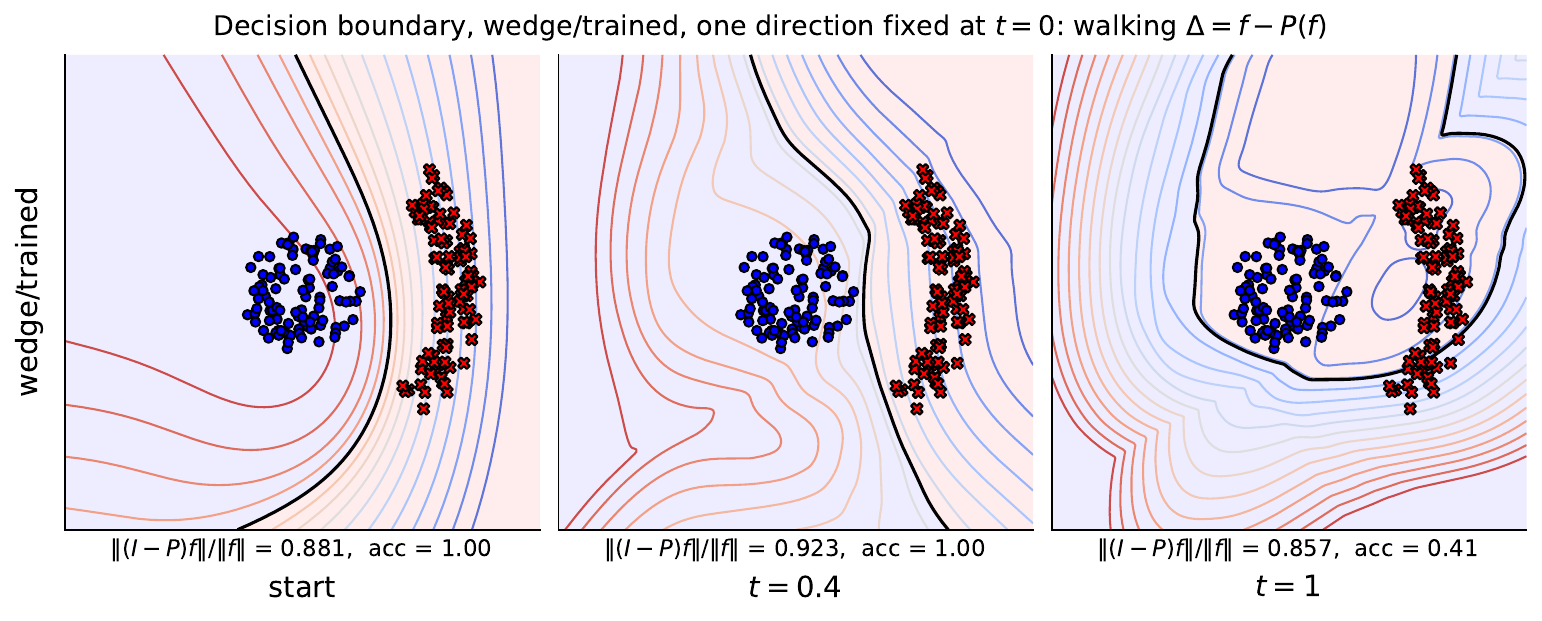}

        \smallskip
        {\small (a) Invariance-directed flow}
    \end{minipage}
    \hfill
    \begin{minipage}[t]{0.48\linewidth}
        \centering
        \includegraphics[width=\linewidth]
        {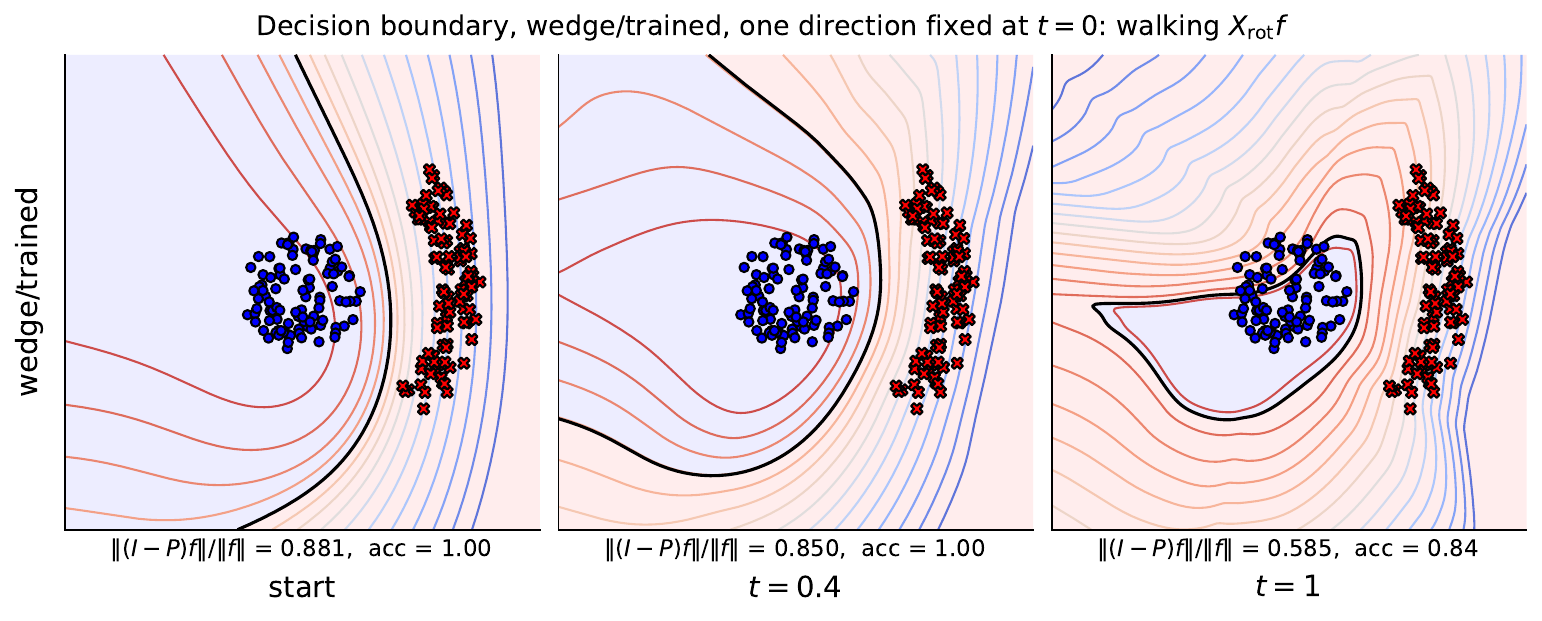}

        \smallskip
        {\small (b) Symmetry-orbit flow}
    \end{minipage}

    \caption{
    Fixed parameter-space flows for the wedge-trained classifier. (a) The initial equivariance-directed direction $b^\star$ does not continue to move the realised function towards the equivariant subspace.
    (b) The initial symmetry-directed direction $c^\star$ deviates from the intended rotational symmetry orbit.
    In both cases, the direction is computed once at $t=0$ and kept fixed throughout the trajectory.
    }
    \label{fig:annulus_fixed}
\end{figure}

\end{document}